\documentclass{article} 
\usepackage{iclr2023_conference,times}

\usepackage{amsmath,amsfonts,bm}

\def\eqref#1{equation~\ref{#1}}

\def\1{\bm{1}}

\DeclareMathAlphabet{\mathsfit}{\encodingdefault}{\sfdefault}{m}{sl}
\SetMathAlphabet{\mathsfit}{bold}{\encodingdefault}{\sfdefault}{bx}{n}

\usepackage{hyperref}
\usepackage{url}
\usepackage{algorithm}
\usepackage{algorithmic}
\usepackage{pgfplots}
\usepackage{graphicx}

\usepackage{eqparbox}

\title{Siamese-NAS: Using Trained Samples Efficiently to Find Lightweight Neural Architecture by Prior Knowledge}

\author{Yu-Ming Zhang  \\
Department of Computer Science \\
National Central University \\
Taoyuan, Taiwan \\
\texttt{nk108522036@g.ncu.edu.tw} \\
\And
Jun-Wei Hsieh \\
College of AI and Green Energy \\
National Yang Ming Chiao Tung University\\
Hsinchu, Taiwan \\
\And
Chun-Chieh Lee \\
Department of Computer Science \\
National Central University \\
Taoyuan, Taiwan \\
\And
Kuo-Chin Fan \\
Department of Computer Science \\
National Central University \\
Taoyuan, Taiwan \\
}

\newcommand{\myeqref}[1]{Eq. (\ref{#1})}
\newcommand{\myfigref}[1]{Fig. \ref{#1}}
\newcommand{\mytabref}[1]{Table. \ref{#1}}
\newcommand{\myalgref}[1]{Alg. \ref{#1}}

\begin{document}
\maketitle

\begin{abstract}
In the past decade, many architectures of convolution neural networks were designed by handcraft, such as Vgg16, ResNet, DenseNet, etc. They all achieve state-of-the-art level on different tasks in their time. However, it still relies on human intuition and experience, and it also takes so much time consumption for trial and error. Neural Architecture Search (NAS) focused on this issue. In recent works, the Neural Predictor has significantly improved with few training architectures as training samples. However, the sampling efficiency is already considerable. In this paper, our proposed Siamese-Predictor is inspired by past works of predictor-based NAS. It is constructed with the proposed Estimation Code, which is the prior knowledge about the training procedure. The proposed Siamese-Predictor gets significant benefits from this idea. This idea causes it to surpass the current SOTA predictor on NASBench-201. In order to explore the impact of the Estimation Code, we analyze the relationship between it and accuracy. We also propose the search space Tiny-NanoBench for lightweight CNN architecture. This well-designed search space is easier to find better architecture with few FLOPs than NASBench-201. In summary, the proposed Siamese-Predictor is a predictor-based NAS. It achieves the SOTA level, especially with limited computation budgets. It applied to the proposed Tiny-NanoBench can just use a few trained samples to find extremely lightweight CNN architecture.
\end{abstract}

\section{Introduction}
There are lots of Convolution Neural Network (CNN) models have been proposed that achieved great success in the past decade (\cite{vgg16,resnet,densnet,mobilenet}). Nevertheless, designing a handcrafted CNN architecture requires human intuition and experience, which is not easy to get and often not optimal. Neural Architecture Search (NAS) (\cite{nas}) focus on this problem, it search the best neural network architecture based on specific strategy on the specific search space (\cite{darts,NASBench101,NASBench201}). Many methods have been proposed in the past. One path is to train a predictor based on Graph Convolution or Multilayer Perceptron. Although the predictor-based method achieved impressive performance with a few trained samples that are randomly CNN architecture trained on CIFAR-10 (\cite{cifar}) or other public datasets. However, to get the ground truth of architectures, we have to train every architecture, which needs so many GPU times, and we all know that GPU times are so expensive. In other words, we believe there is still a huge improvement gap. The proposed Siamese-Predictor combined the extra prior knowledge Estimation Code to improve the search efficiency. The proposed Estimation Code is developed from early losses of the architecture training procedure. It shows a high correlation with the final accuracy of the architecture in our experiments. Although this code shows the potential to boost the predictor, it has a huge computational cost for getting the Estimation Code of each model in the search space. So, we propose the Siamese-Ranking method to reduce this extra cost brought by extra knowledge. Besides, we also propose the training strategy Batch Top Sampling to use trained samples more efficiently by conditional training sample selection. Finally, our method surpasses the baseline on the NASBench-201 (\cite{NASBench201}) and finds more lightweight CNN architecture on the proposed Tiny-NanoBench.

\section{Related Work}
There are various NAS methods have been proposed in latter years, some methods based on reinforcement learning (\cite{nas,dsgnasrl}), some methods developed from evolutionary algorithm (\cite{nsganet,reicnas}), and some methods focused on training a predictor to sort the search space (\cite{nrlprdctr,brpnas}). In this papaer, we proposes the Siamese-Predictor, it is a predictor-based NAS method. Therefore, We will focus on introducing the other predictors later.
\subsection{Neural Predictors}
The predictor-based NAS methods are the mainstream approach, creating a predictor to predict the accuracy of CNN architecture. There are many types of these predictors (\cite{pnas,nao,nrlprdctr,brpnas,npenas}), in recent works, the Neural Predictor (\cite{nrlprdctr}) is the most representative method. It encodes the architecture to graph structure embedded code that contains the adjacency matrix and feature matrix. The adjacency matrix represents the relationship between nodes. The feature matrix represents the operations used by the nodes in the cell. Since the number of cells in their meta-architecture is fixed and the structure is also fixed. So that uses multiple Graph Convolution (GCN) (\cite{gcn}) to extract the high-level feature matrix while considering directionality by the adjacency matrix. Finally, flatten it and use a fully connected layer to get the prediction accuracy, then sort search space based on it to find the promising architectures. This GCN-based predictor shows significant performance in NASBench-101 (\cite{NASBench101}). After that, the BRP-NAS (\cite{brpnas}) proposes a binary predictor that takes two different architectures as inputs at the same time, and it predicts which is better in these two instead of directly predicting accuracy. This method greatly improves the performance compared to the neural predictor. We believe that this improvement is caused by the use of prior knowledge, i.e., inputs two architectures at the same time. In this paper, we explore this idea to propose the Siamese-Predictor that using losses of training procedure as prior knowledge, it improves the performance considerably in NASBench-201 (\cite{NASBench201}).

\section{Methods}
\begin{figure}
\centering
\begin{tabular}{ccc}
\fbox{\includegraphics[width=4cm]{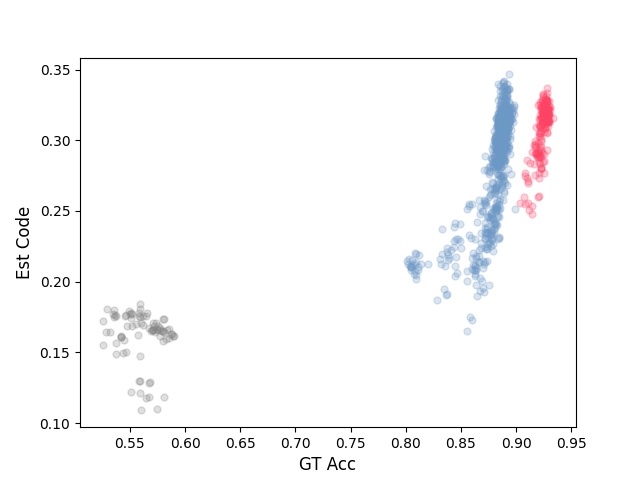}}&
\fbox{\includegraphics[width=4cm]{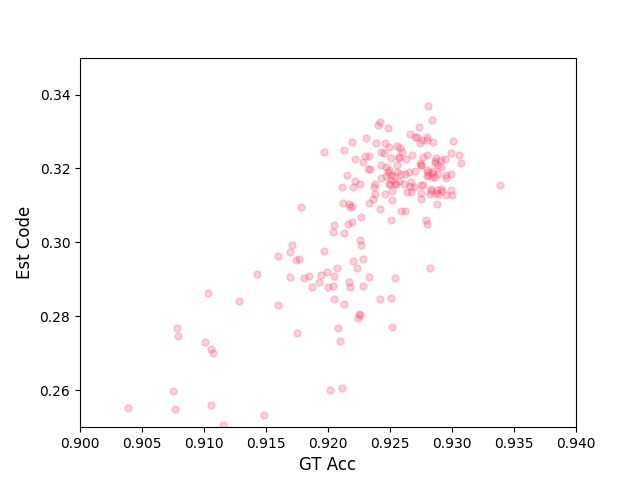}}&
\fbox{\includegraphics[width=4cm]{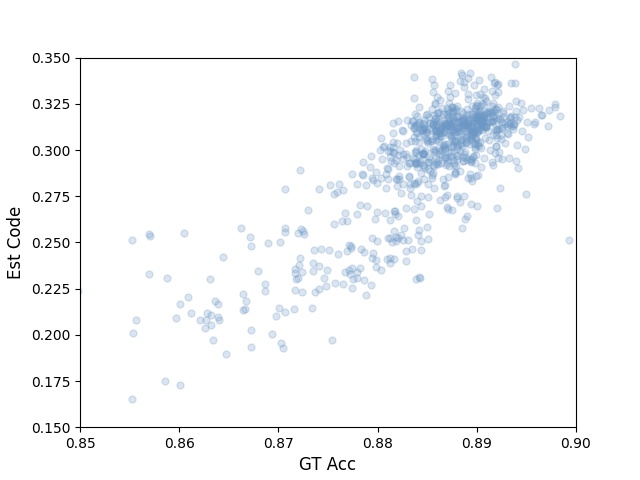}}\\
(a)&(b)&(c)
\end{tabular}
\caption{The relationship between Estimation Code and ground truth(accuracy) in Tiny-NanoBench. (a) is the relationship about all architecture, (b) is focused on the first high-scoring group in (a), (c) is focused on the second high-scoring group in (a).}
\label{estcodeanalyst}
\end{figure}

This section introduces the proposed Estimation Code, Siamese-Predictor, and Siamese-Ranking. The Estimation Code is prior knowledge about the final accuracy of architecture based on training procedure. We believe that the performance of the predictor will be improved by combining prior knowledge. It inspired the past work (\cite{brpnas}). They used two different architectures as inputs at the same time. They think doing this will give the predictor more information. Our observation \myfigref{estcodeanalyst} also shows a high correlation between Estimation Code and accuracy of architecture. For the proposed Siamese-Predictor, we mixed naive predictor and Estimation Code. It consists of two parts, the basic branch and the estimation branch, the consumption of the two branches in the prediction stage is inconsistent. So, the Siamese-Ranking is about how to use them efficiently to predict every accuracy of search space.

\begin{algorithm}[th]
\caption{Siamese Ranking}
\label{simsrank}
\textbf{Input}: $(i)$ search sapce $\mathcal S$, $(ii)$ basic branch of siamese-predictor $P_b$, $(iii)$ estimation branch of siamese-predictor $P_e$.\\
\textbf{Output}: ranked search sapce $\mathcal R$.
\begin{algorithmic}[1] 
\STATE $\mathcal R_b \gets \emptyset$ 
\STATE $\mathcal R_e \gets \emptyset$
\WHILE{$s \in \mathcal S$}
\STATE $s_{adj}, s_{feat} \gets s$ \COMMENT{getting adjacency matrix and feature matrix.} 
\STATE $p_b \gets P_b(s_{adj}, s_{feat})$ \COMMENT{getting prediction by basic branch.} 
\STATE insert $s$ to $\mathcal R_b$ based on $p_b$. 
\ENDWHILE
\STATE $\mathcal R_{c} \gets Top(\mathcal R_b,c)$ \COMMENT{take out the top $c$.} 
\WHILE{$s \in \mathcal R_{c}$}
\STATE $s_{adj}, s_{feat} \gets s$
\STATE $s_{estc} \gets EstCode(s_{adj}, s_{feat})$ \COMMENT{for $3$ epochs(0.3\% cost of complete training).} 
\STATE $p_e \gets P_e(s_{adj}, s_{feat}, s_{estc})$ \COMMENT{getting prediction by estimaion branch.} 
\STATE insert $s$ to $\mathcal R_e$ using $p_e$. 
\ENDWHILE
\STATE $\mathcal R \gets$ replace $\mathcal R_c \cap \mathcal R_b$ with $\mathcal R_e$. \COMMENT{this means resorting the top $c$.} 
\STATE \textbf{return} $\mathcal R$
\end{algorithmic}
\end{algorithm}

\subsection{Estimation Code}
The proposed Estimation Code is developed from the loss of model training procedure. It is an intuitive idea about the relationship between the loss value and the ground truth accuracy of an architecture. Based on this idea, We used the first three losses as the Estimation Code. In order to explore and evaluate this naive idea, we are experimenting with the Tiny-NanoBench we built. The \myfigref{estcodeanalyst} shows that the Estimation Code has a local positive correlation with accuracy, especially in the high-accuracy group. However, to get an architecture's Estimation Code (first three losses), we have to train this architecture on CIFAR-10 or the other dataset for three epochs. Considering that the whole search space contains thousands of architectures, this will cause much computational cost in the search stage. Therefore, we construct the mini-batch of CIFAR-10 by random sampling with a one-tenth ratio, then train from scratch on this mini-batch for three epochs to get the Estimation Code. In this way, the consumption of computing resources can be significantly reduced. So the consumption of Estimation Code compared to the entire training is 50,000 times 100 epochs and 5,000 times three epochs. The consumption is only 0.3\% of the complete training process for an architecture, which is almost negligible. \myeqref{estcodeidea} shows that the idea of the proposed Estimation Code, $g(M)$ denotes the ground truth of architectures, $p(M)$ denotes the prediction accuracy of architectures by predictor, $\theta$ denotes the weights of predictor, $I$ denotes the prior knowledge (Estimation Code), $S_0,S_1,S_2$ denotes the budget using of trained samples for predictor or Estimation Code. Therefore, we believe that the prior knowledge brought by Estimation Code can make the Siamese-Predictor more accurate while $S_1+S_2$ is more minor than $S_0$.

\begin{equation}
\begin{aligned}
min_{(\theta)}\ (g(M)-p(M|\theta))^2&\simeq(g(M)-p(M|S_0))^2,\\
min_{(\theta,I)}\ (g(M)-p(M|\theta,I))^2&\simeq(g(M)-p(M|S_1,S_2))^2,\\
then\ (S_1+&S_2)<S_0
\end{aligned}
\label{estcodeidea}
\end{equation}

\subsection{Siamese-Predictor}
This section continues the previous section. We will describe how to integrate the proposed Estimation Code into the native predictor. In general, adding some extra prior knowledge to the predictor may boost the performance. But as mentioned in \myeqref{estcodeidea}, it is important to properly control the consumption of budgets $S_1+S_2$ to be less than $S_0$. This makes us build a predictor with two modes. The first mode only uses the adjacency matrix and feature matrix to predict the accuracy of architecture. The second mode uses these and the Estimation Code as the third input. So far, the proposed Siamese-Predictor consists of two branches. The \myfigref{siamesepredictor} shows that the left branch represents the first mode. We call it the basic branch. The right branch represents the second mode. We call it the estimation branch, and as we say, the estimation branch predicts accuracy involving prior knowledge. It should be better than the basic branch. In the fusion process of the 2-d feature matrix and 1-d Estimation Code, we are first upsampling the Estimation Code to the 2-d matrix by a fully-connected layer and reshaping operation. Then the feature matrix and upsampled Estimation Code matrix through the proposed Estimation Fusion Module fuse the feature and prior knowledge. Next, we will introduce the structure of EFM.

\subsubsection{Estimation Fusion Module (EFM)}
The self-Attention has been widely used in vision CNN models in recent years (\cite{vit,swntrnsfmr}), so we think the combination of self-Attention and Graph Convolution is worth trying. The self-Attention split an input as three different - q, k, v by linear or non-linear transformation, then fused q and k to find a self-correlation, and through softmax to factorization. Finally, it multiplied to v to act on itself. In our case, The proposed estimation fusion module (EFM) is a kind of self-Attention method. Actually, EFM is a cross-Attention method between the feature matrix and Estimation Code. We regard the feature matrix as q,v by Graph Convolution, then treat the upsampled Estimation Code as k by Graph Convolution. However, a predictor for NAS and a classifier for classification tasks are not the same, and we have to consider the direction property of the Directed Acyclic Graph (DAG) for the predictor of NAS. So, after we have calculated q multiplied by k, we will multiply the adjacency matrix to prove the direction of information flow. Finally, the skip connects are attached to the transformed feature matrix to enhance the feature passing. Finally, it goes through a Graph Convolution. The part (a) of \myfigref{efmnam} shows the structure of the EFM. The part (a) of \myfigref{siamesepredictor} shows where EFM was attached.

\subsubsection{Nodes self-Attention Module (NSAM)}
\begin{figure}[t]
\centering
\begin{tabular}{cc}
\fbox{\includegraphics[width=5.35cm]{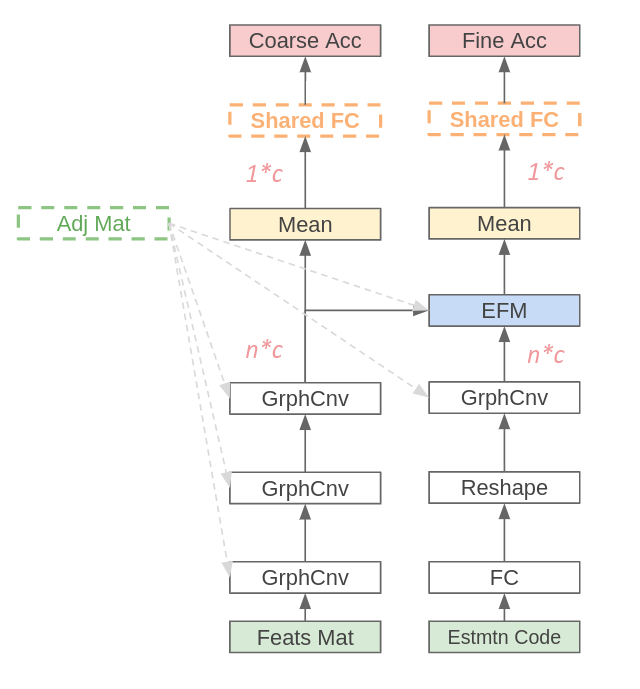}}&
\fbox{\includegraphics[width=5.35cm]{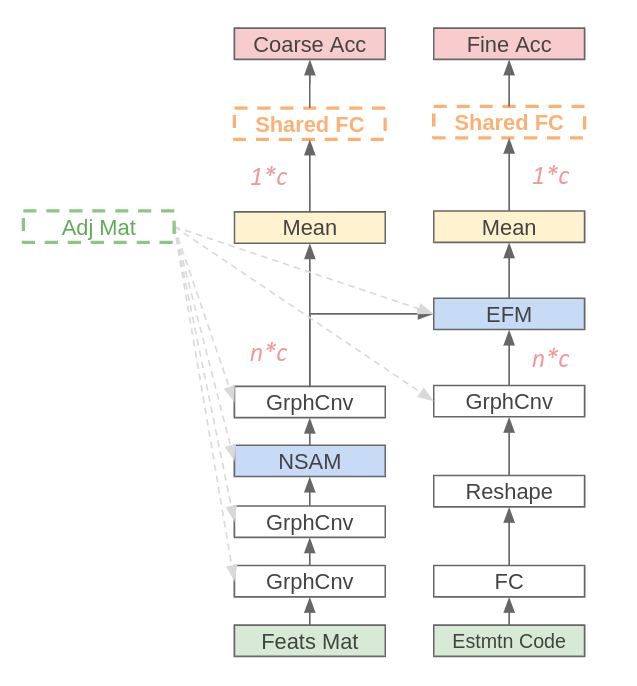}}\\
(a)&(b)
\end{tabular}
\caption{The structure of Siamese-Predictor. Figure (a) shows the Siamese-Predictor. The basic branch only uses the adjacency matrix and feature matrix, and the estimation branch uses an additional Estimation Code as another input. Figure (b) shows the Siamese-Predictor with NSAM. The red character "n" means the number of nodes, and the "c" means the feature length that is used to represent the node.}
\label{siamesepredictor}
\end{figure}

\begin{figure}[t]
\centering
\begin{tabular}{cc}
\fbox{\includegraphics[width=5.8cm]{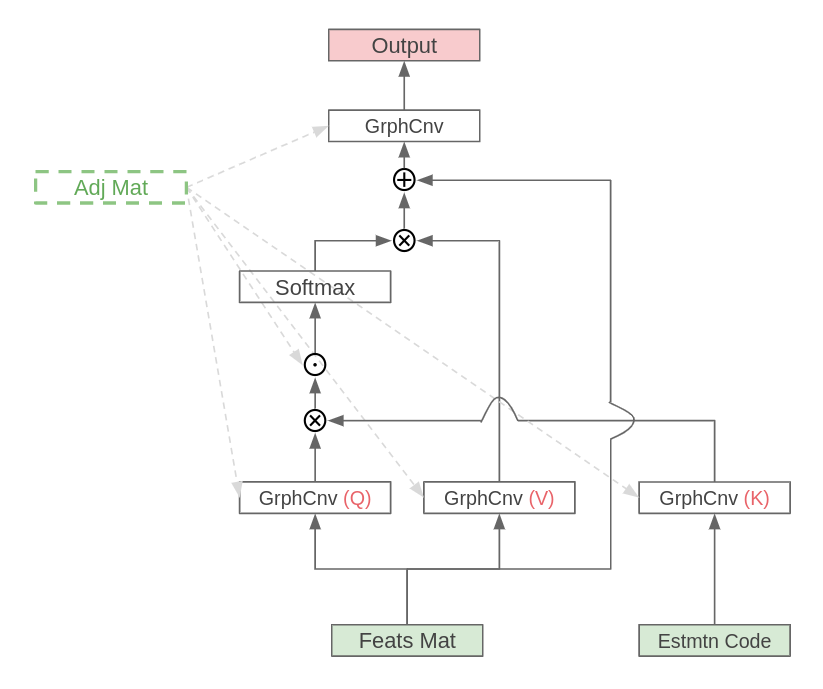}}&
\fbox{\includegraphics[width=6.1cm]{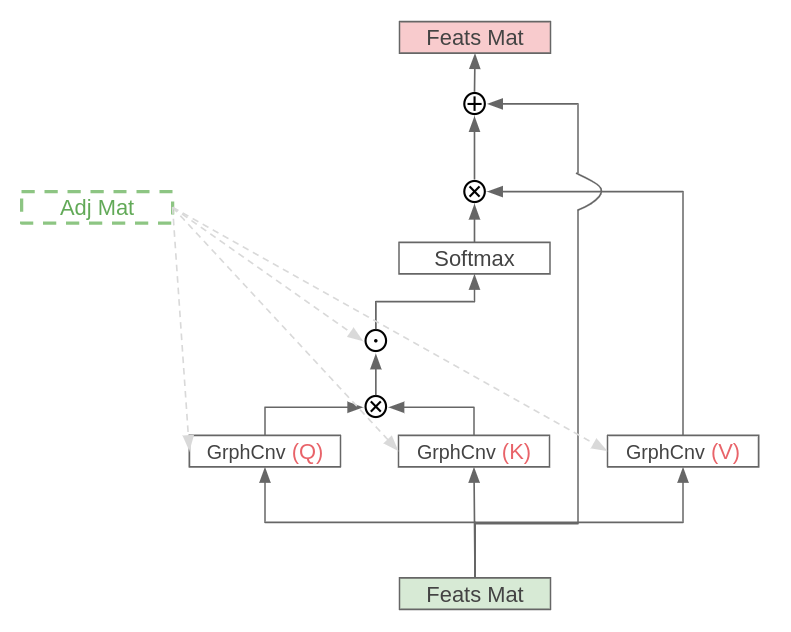}}\\
(a)&(b)
\end{tabular}
\caption{Figure (a) shows the structure of EFM. Figure (b) shows the the structure of NSAM. The symbol $\bigotimes$ means the operation of matrix multiplication, and the symbol $\bigodot$ means the operation of Hadamard product, the symbol $\bigoplus$ just means the element-wise addition of two matrixes.}
\label{efmnam}
\end{figure}

Unlike the estimation fusion module, the proposed Nodes self-Attention Module(NSAM) is a pure self-Attention method with Graph Convolution focused on the relation of the same input. It used Graph Convolution to generate the q, k, and v matrix from the same input feature matrix. Then we also multiply the adjacency matrix as EFM for the same reason. Part (b) of \myfigref{efmnam} shows the structure of the NSAM. In later experiments, we will show more comparisons about NSAM. Part (b) of \myfigref{siamesepredictor} shows where NSAM was attached.

\begin{algorithm}[t]
\caption{Batch Top Sampling}
\label{algbts}
\textbf{Input}: $(i)$ search sapce $\mathcal S$, $(ii)$ training samples pool $\mathcal T$, $(iii)$ batch size $b$, $(iv)$ max iteration $l$, $(v)$ training samples budgets $m$, $(vi)$ update frequency $f$ (default is $10$).\\
\textbf{Output}: siamese predictor $P$.
\begin{algorithmic}[1] 
\STATE $P \gets Init(\theta)$ \COMMENT{initialize the weights of predictor.} 
\STATE $\mathcal T \gets RndChoices(\mathcal S, \lambda m)$ \COMMENT{initialize the pool, and $\lambda$ is between 0 and 1.} 
\WHILE{$i<\alpha l$}
\STATE $P \gets Grad(P, RndChoices(\mathcal T, b))$ \COMMENT{updating predictor by gradient of batch data.}
\STATE $i \gets i+1$
\ENDWHILE
\WHILE{$i<(1-\alpha)l$}
\IF {$i \% f==0$}
\STATE $\mathcal S_f \gets RndChoices(\mathcal S, Size(S)/f)$ \COMMENT{random subset of search space.}
\STATE $\mathcal R_f \gets SimsRanking(\mathcal S_f)$ \COMMENT{as shown in \myalgref{simsrank}.}
\STATE $\mathcal T \gets \mathcal T \cup Top(\mathcal R_f,(1-\lambda)m/f)$. \COMMENT{this means take top $(1-\lambda)m/f$ into pool.}
\ENDIF
\STATE $P \gets Grad(P, RndChoices(\mathcal T, b))$
\STATE $i \gets i+1$
\ENDWHILE
\STATE \textbf{return} $P$
\end{algorithmic}
\end{algorithm}

\subsection{Siamese-Ranking}
The proposed Siamese-Predictor has two branches: the basic and the estimation branches. The basic branch is used to predict coarse accuracy almost for computation free because no extra trained samples is needed. The estimation branch is used to predict fine accuracy but is more expensive than the basic branch because getting the Estimation Code needs extra trained samples for three epochs. On the other hand, $S_1+S_2$ must be smaller than $S_0$. So, we propose a mixture ranking method that first sorts search space by basic branch for almost computational cost-free, then just sorts top $c$ by estimation branch. It ensures the stability of the high-accuracy group and solves the expensive problem. The \myalgref{simsrank} shows the implementation detail. The top $c$ is an essential parameter of this algorithm, and if $c$ is too large, the cost will be unacceptable. So the $c$ is set to 30 in BTS, and the $c$ is set to 60 in the evaluation stage. The computational cost for getting the Estimation Code of an architecture is just 0.3\% compared to the complete training procedure, so the cost when the $c$ is 30 or 60 is not worth mentioning. So far, The proposed Siamese-Ranking solves the budget problem using estimation branch.

\begin{table}[t]
\caption{Comparison of BTS and FTS for training time used Siamese-Predictor. The update frequency $f$ is fixed at 10, and the training samples budget $c$ is fixed at 100.}
\label{btscomp}
\centering
\begin{tabular}{llll}
\multicolumn{1}{c}{\bf Method} &\multicolumn{1}{c}{\bf Search Space} &\multicolumn{1}{c}{\bf Batch Samples}  &\multicolumn{1}{c}{\bf Training Time(s)}
\\ \hline \\
Fully Top Sampling &Tiny-NanoBench(1120) &1120 &115.67 \\
Batch Top Sampling &Tiny-NanoBench(1120) &112 &\bf 67.22 \\
Fully Top Sampling &NASBench-201(15,625) &15,625 &816.43 \\
Batch Top Sampling &NASBench-201(15,625) &1563 &\bf 140.62
\end{tabular}
\end{table}

\subsection{Batch Top Sampling (BTS)}
The previous sections introduced how to build a powerful predictor with extra prior knowledge Estimation Code. In addition, a sampling strategy is also essential for predictor-based NAS, and a good strategy can train a good predictor using a few trained samples. The proposed Batch Top Sampling (BTS) randomly selects a few training samples to the training pool, then iteratively performs the following steps: 1) Randomly selects some trained samples to train the temporary Siamese-Predictor from the training pool. 2) Randomly selects some samples to form a batch search space from the whole search space. 3) Sorting (Siamese-Ranking) the batch search space based on the temporary Siamese-Predictor. 4) Choosing the top $n$ samples of batch search space to the training pool. The \myalgref{algbts} shows the implementation detail. This strategy of coarse to fine during training keeps the predictor focused on the high-accuracy groups. However, it also consumes considerable computing resources to predict architecture in the whole search space. The proposed BTS randomly selected to form the batch search space at the first step solved this problem. The \mytabref{btscomp} shows that the training time of BTS is faster than Fully Top Sampling(FTS) due to the BTS using batch search space and FTS using whole search space. Finally, the BTS can help the predictor to boost the performance and reduce the variance from training randomness in later experiments.

\begin{figure}[t]
\centering
\begin{tabular}{cc}
\fbox{\includegraphics[width=6cm]{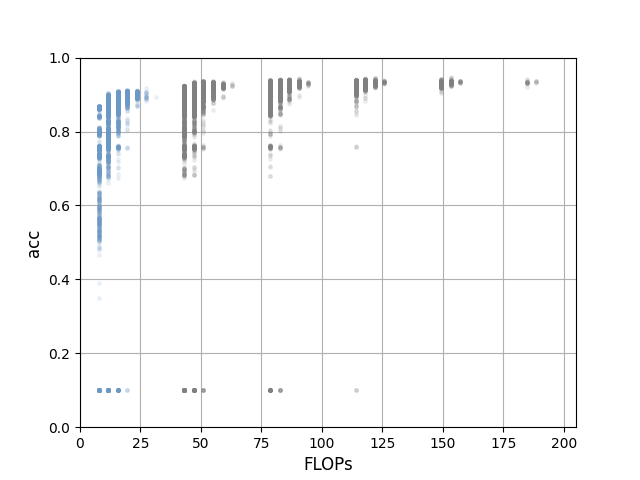}}&
\fbox{\includegraphics[width=6cm]{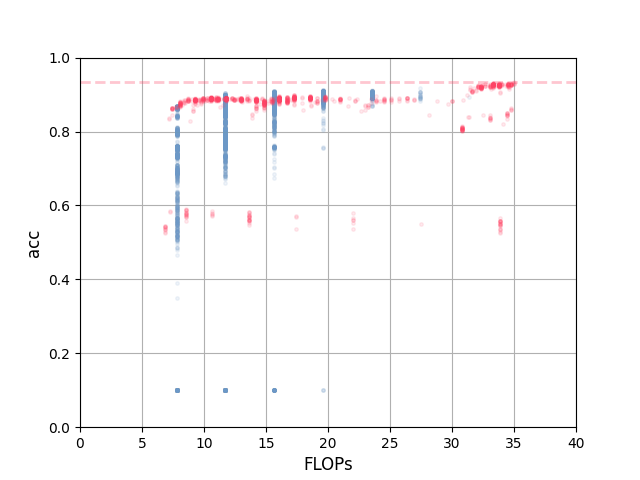}}\\
(a)&(b)
\end{tabular}
\caption{Figure (a) is the distribution of NASBench-201, and figure (b) is the distribution comparison of NASBench-201 (blue) and the proposed Tiny-NanoBench (red).}
\label{cmpdstrbt}
\end{figure}

\subsection{Well-Designed Lightweight Search Space}
In the previous sections, we proposed Siamese-Predictor, components, and tricks. Although this robust NAS algorithm shows considerable performance in later experiments. Our another purpose is focused on lightweight architecture search. As part (a) of \myfigref{cmpdstrbt} shown, we record the relationship between FLOPs and accuracy for NASBench-201. We can find that it consists of several groups at the FLOPs axis. Based on this observation, we set a constraint of about 35 MFLOPs to focus on the smallest FLOPs group (blue group in part (a) of \myfigref{cmpdstrbt}), we call this subset Tiny-NASBench-201. We believe that Tiny-NASBench-201 is not the optimal search space at the tiny FLOPs (~35M) level, so we proposed the Tiny-NanoBench trained on CIFAR-10. It is a well-designed lightweight search space based on a series of lightweight operations, depthwise convolution, pointwise convolution (\cite{mobnetv2}), splitting on the channel (\cite{cspnet}), expanding channel block (\cite{cslyolo}), ......, etc. We also record the relationship between the FLOPs and accuracy. We found that the distribution of Tiny-NanoBench is more stable than Tiny-NASBench-201, especially at 10~15 MFLOPs. We also found that the Tiny-NanoBench has better architectures than Tiny-NASBench-201 at 30~35 MFLOPs.

\section{Experiments}
In this section, we conduct a series of experiments to validate our proposed methods on two benchmarks. We introduced the public search space NASBench-201 and evaluated the proposed Siamese-Predictor on it. Then we found that the typical training budget of $N+K$ is something worth thinking about the trade-off. Besides, We also introduced the proposed search space Tiny-NanoBench and evaluated Siamese-Predictor on it for lightweight architecture search.

\subsection{General Evaluation on NASBench-201}
\begin{figure}[t]
\centering
\begin{tabular}{cc}
\fbox{\includegraphics[width=6cm]{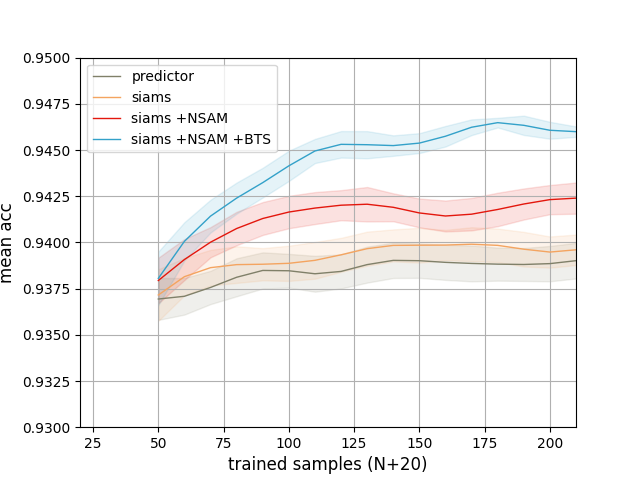}}&
\fbox{\includegraphics[width=6cm]{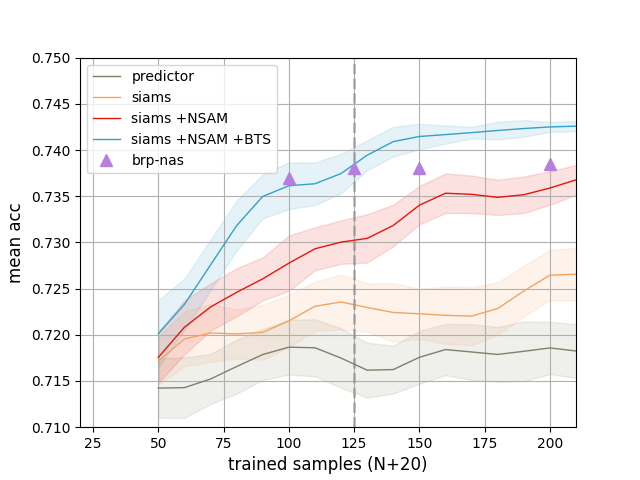}}\\
(a)&(b)
\end{tabular}
\caption{The experiment of Siamese-Predictor on NASBench-201. Figure (a) shows the results of CIFAR-10, and figure (b) shows the results of CIFAR-100. The larger the area of the light-colored part, which means the more extensive the standard deviation.}
\label{simsponnas201}
\end{figure}

NASBench-201 is a cell-based search space, it is public, credible, and easy to use, it records the trained accuracy about three dataset - CIFAR-10, CIFAR-100 (\cite{cifar}), ImageNet-16-120 (\cite{imagenet16}). NASBench-201 contains 15,625 different architectures. These architectures are built on the same meta-architecture by different cells with six operations. We put our Siamese-Predictor up for comparison with the naive predictor. The \myfigref{simsponnas201} shows the 100 run results of the comparison. The "trained samples" represent the number of trained samples from the search space used by the predictor during the training process. It also contains the top-$k$ choices at evaluating stage. The "mean acc" represents the average best accuracy of 100 runs. Overall, we trained a predictor with part of "trained samples" and sorted search space by this predictor. Finally, we took the top 20 and trained them one by one from scratch to get the architecture with the highest ground truth for each run. As shown in the \myfigref{simsponnas201}, our method surpasses the naive predictor in the case of CIFAR-10, especially in CIFAR-100, the performance gap is greater than CIFAR-10. We also found that the NSAM and BTS bring considerable improvements in these two dataset. The proposed Siamese-Predictor with NSAM and BTS surpasses the SOTA brp-nas when trained samples are greater than 125, and the 125 samples only account for 0.8\%(125/15265) of NASBench-201, even using 200 trained samples, that is only 1.3\%, and it still achieves SOTA level with such few budgets. When trained samples are about 200, we also found that the standard deviation of Siamese-Predictor +NSAM +BTS is more stable than the other version. We believe this is due to BTS picking samples more efficiently.

\begin{figure}[t]
\centering
\begin{tabular}{cc}
\fbox{\includegraphics[width=6cm]{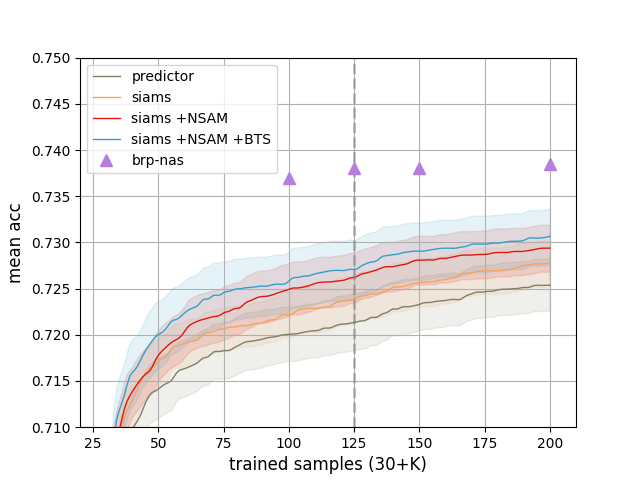}}&
\fbox{\includegraphics[width=6cm]{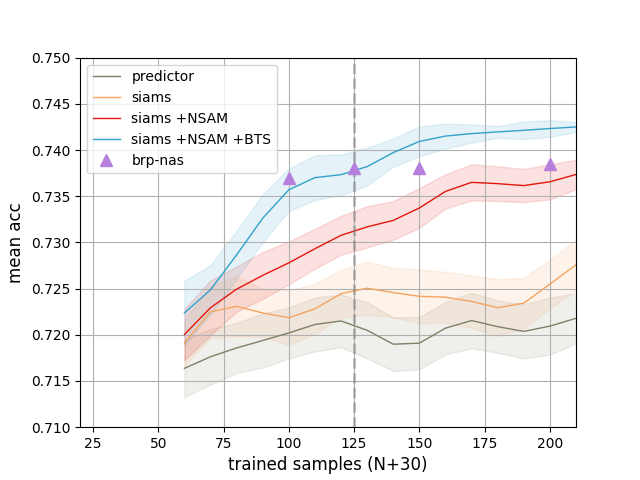}}\\
(a)&(b)
\end{tabular}
\caption{Comparison of the trade-off between $N$ and $K$. Figure (a) is fixed $N$ and increase $K$, and figure (b) is to fix $K$ then increase $N$.}
\label{cmpnk}
\end{figure}

\subsubsection{Trade-off About $N+K$}
The "trained samples" is a critical metric for predictor-based NAS. It consists of two variables, the number of training architectures for predictor $N$ and the top $K$ of sorted search space at the evaluation stage. In past, neural predictors (\cite{nrlprdctr,brpnas}) prefer to fixed $N$ and increase the $K$ gradually. Nevertheless, we believe that it is to fix $K$, and increasing $N$ may improve the performance because the number of $N$ directly correlates with training samples for predictor training. The \myfigref{cmpnk} shows the experiments on NASBench-201, which is fixed $N$ or $K$ and evaluates these predictors. Part (a) of \myfigref{cmpnk} is fixed $N$ to 30, then gradually increases $K$ to draw the entire curve. We can see that when $N$ even increased to 170 (the sum is 200), it is still far away from SOTA. On the other hand, part (b) of \myfigref{cmpnk} is fixed $K$ to 30, then gradually increases $N$. We found that spending the bulk of the budget to $N$ can significantly improve. Not only that, it can be seen that as $N$ increases, the standard deviation also decreases, and increasing $K$ has no noticeable effect. These obvious conditions lead us to believe that increasing $N$ is a better choice.

\begin{figure}[t]
\centering
\begin{tabular}{cc}
\fbox{\includegraphics[width=6cm]{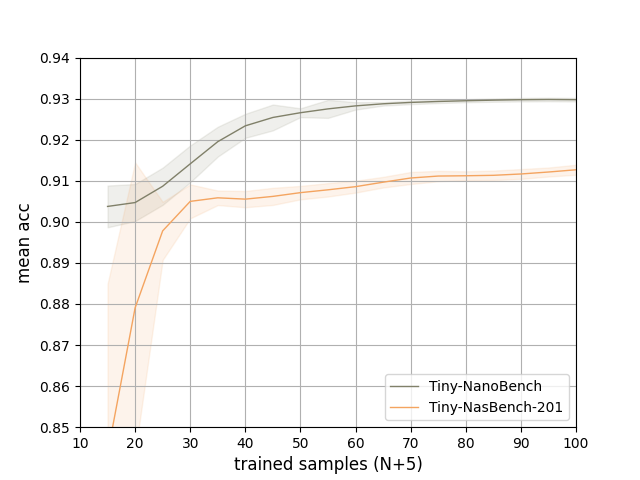}}&
\fbox{\includegraphics[width=6cm]{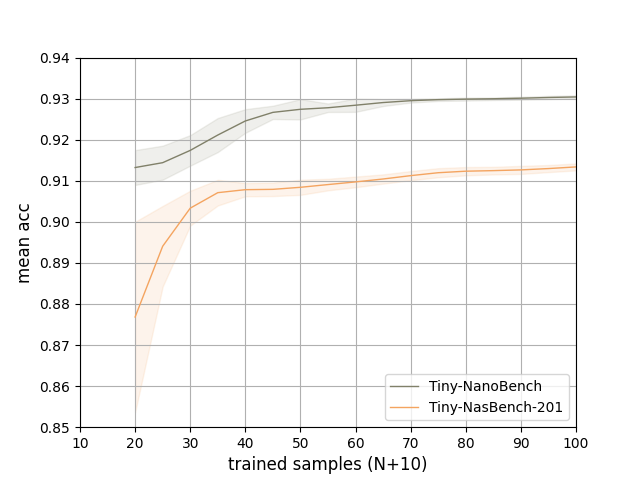}}\\
(a)&(b)
\end{tabular}
\caption{The comparison of lightweight searching is between the Tiny-NanoBench and the Tiny-NASBench-201. Figure (a) is to fix $K$ as 5, and figure (b) is to fix $K$ as 10.}
\label{cmpNASBench}
\end{figure}

\subsection{Lightweight Evaluation on Tiny-NanoBench}
In order to verify the benefit of the proposed search space Tiny-NanoBench. We extract a small subset Tiny-NASBench-201 from NASBench-201. The exact extraction method is described in section 3. Both search spaces have the same constraints on FLOPs ($<35M$). We use the proposed Siamese-Predictor to find the best architecture on these two search spaces, as shown in part (a) of \myfigref{cmpNASBench}. We found that it finds the better architecture faster under Tiny-NanoBench, especially when $K$ and $N$ are small. Part (a) and (b) of \myfigref{cmpNASBench} show that it also can find the architecture of higher accuracy than Tiny-NASBench-201 when $N$ is large.

\begin{figure}[t]
\centering
\begin{tabular}{cccc}
\fbox{\includegraphics[width=2.8cm]{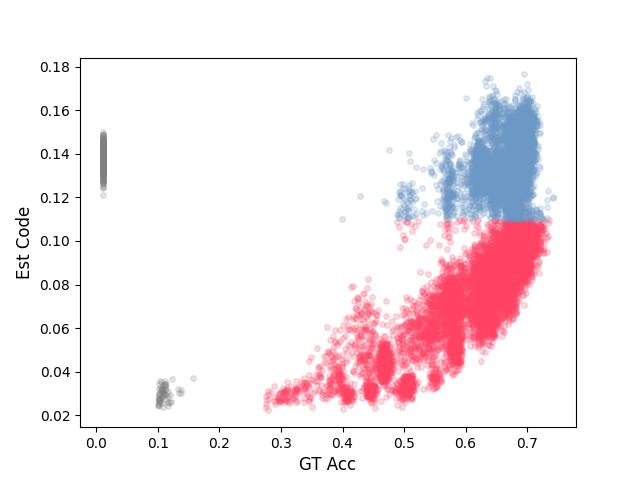}}&
\fbox{\includegraphics[width=2.8cm]{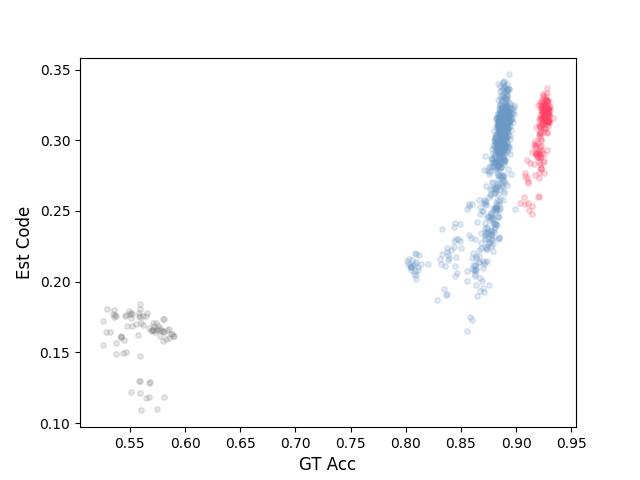}}&
\fbox{\includegraphics[width=2.8cm]{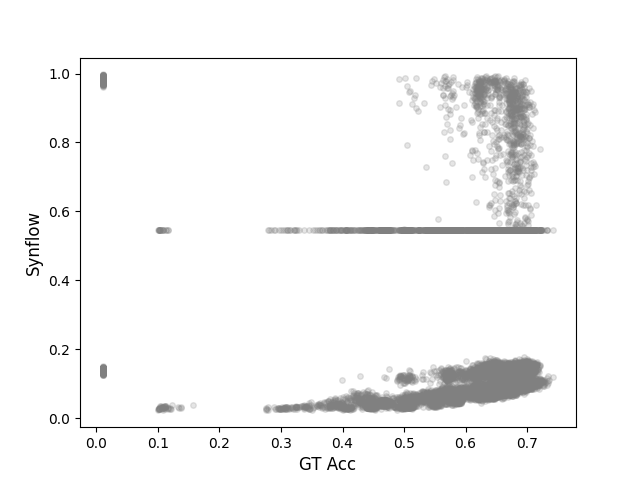}}&
\fbox{\includegraphics[width=2.8cm]{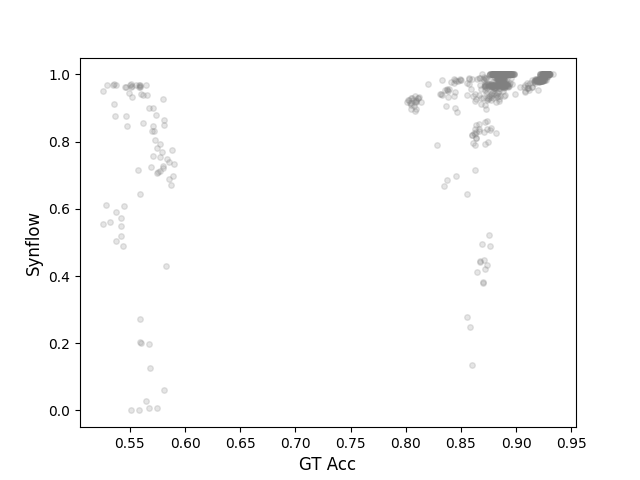}}\\
(a)&(b)&(c)&(d)
\end{tabular}
\caption{The comparison of Estimation Code and synflow. Figure (a) is the distribution of NasBench-201 (CIFAR-100). Figure (b) is the distribution of Tiny-NanoBench. Figure (c) is the distribution of NASBench-201 about accuracy and synflow. Figure (d) is the distribution of Tiny-NanoBench about accuracy and synflow.}
\label{cmpestcode}
\end{figure}

\subsection{Analysis for Estimation Code}
The proposed Estimation Code is the critical part of this paper. In order to further analyze it, we introduce the other cost-free metric of prior knowledge. This work (\cite{zerocostnas})detailed introduces several metrics. There are snip (\cite{snip}), grasp (\cite{grasp}) and synflow ,and the synflow (\cite{synflow}) shows the considerable performance for many NAS methods in their experiments. So, we take the synflow to replace the Estimation Code as the prior knowledge for comparison. The \myfigref{cmpestcode} shows the results on NASBench-201 and Tiny-NanoBench. The red and blue regions of \myfigref{cmpestcode} have a significant positive correlation between prior knowledge (estimated code or synchronous flow) and accuracy. The gray region is the opposite. So, we can see that part (c) and part (d) of \myfigref{cmpestcode} are more disorders than part (a) and part (b) of \myfigref{cmpestcode}, which means the Estimation Code has a higher correlation with accuracy than synflow.
 
\section{Conclusion}
This paper focuses on building a powerful predictor to find promising CNN architecture for training samples efficiently. The proposed Siamese-Predictor using EFM to fuse prior knowledge - Estimation Code. The proposed Estimation Code and accuracy of architectures are highly correlated based on our observation (\myfigref{estcodeanalyst}). Besides, we also proposed Graph Convolution-based self-Attention method NSAM and fast and efficient sampling strategy BTS. These two tricks greatly enhanced the performance of the Siamese-Predictor in our experiments. It surpass BRP-NAS and achives the SOTA level on NASBench-201 (\myfigref{simsponnas201}). We also proposed a search space Tiny-NanoBench, for lightweight CNN architectures searching. In this search space, our predictor can find better lightweight architecture more easily than Tiny-NASBench-201, the lightweight subset of NASBench-201. In summary, we proposed a powerful and efficient predictor-based NAS method Siamese-Predictor and a search space Tiny-NanoBench designed for lightweight CNN architecture.

\newpage
\bibliography{iclr2023_conference}
\bibliographystyle{iclr2023_conference}




\end{document}